# Automatically Explaining Machine Learning Prediction Results: A Demonstration on Type 2 Diabetes Risk Prediction

**Gang Luo** (corresponding author)

Department of Biomedical Informatics, University of Utah, Suite 140, 421 Wakara Way, Salt Lake City, UT 84108, USA

gang.luo@utah.edu

Phone: 1-801-213-3565

**Abstract**

**Background:** Predictive modeling is a key component of solutions to many healthcare problems. Among all predictive modeling approaches, machine learning methods often achieve the highest prediction accuracy, but suffer from a long-standing open problem precluding their widespread use in healthcare. Most machine learning models give no explanation for their prediction results, whereas interpretability is essential for a predictive model to be adopted in typical healthcare settings.

**Methods:** This paper presents the first complete method for automatically explaining results for any machine learning predictive model without degrading accuracy. We did a computer coding implementation of the method. Using the electronic medical record data set from the Practice Fusion diabetes classification competition containing patient records from all 50 states in the United States, we demonstrated the method on predicting type 2 diabetes diagnosis within the next year.

**Results:** For the champion machine learning model of the competition, our method explained prediction results for 87.4% of patients who were correctly predicted by the model to have type 2 diabetes diagnosis within the next year.

**Conclusions:** Our demonstration showed the feasibility of automatically explaining results for any machine learning predictive model without degrading accuracy.

**Background**

Predictive modeling is widely used in many healthcare applications [1]. It is a key technology to transform biomedical data sets into actionable knowledge, advance clinical research, and improve healthcare. For example, predictive models can be used to forecast a patient's health risk, outcomes, or future behavior to provide timely and appropriate care, such as:

(1) Predict whether a diabetes patient will be admitted to the hospital next year. Sign up patients at high risk for hospital admission for a diabetes case management program [2].

(2) Predict an asthmatic patient's total healthcare cost within the next year. Sign up patients at high risk for incurring high costs for an asthma case management program [2].

(3) Predict whether a patient will have type 2 diabetes diagnosis within the next year. Give preventive interventions to patients at high risk for type 2 diabetes [3-7].

Predictive modeling can be conducted through several approaches including using rules developed by medical experts, statistical methods such as logistic regression, and machine learning algorithms that automatically improve themselves through experience [1], such as neural network, Bayes network, support vector machine, decision tree, random forest, and boosted regression tree. Among these approaches, machine learning methods often achieve the highest prediction accuracy [8]. In fact, with less strict assumptions on data distribution, machine learning can sometimes double the prediction accuracy when compared to statistical methods [8-10].

Despite its superior performance, machine learning is far from being as widely used in healthcare applications as it should be because it suffers from a long-standing open problem. Most machine learning models are complex and give no explanation for their prediction results. Nevertheless, interpretability is essential for a predictive model to be adopted in typical healthcare settings. When lives are at stake, clinicians need to understand the reasons to trust the prediction results. Knowing the reasons for undesirable outcomes can help clinicians choose tailored interventions, which are often more effective than non-specific ones. If sued for malpractice, clinicians need to defend their decisions in court based on their understanding of the prediction results. In addition, explanations for prediction results can give hints to and facilitate discovery of new knowledge in clinical research.

At present, several explanation methods for machine learning predictive models exist, but they are model specific and decrease accuracy [7, 11, 12]. Recently, we proposed the first draft method for automatically explaining results for any machine learning predictive model without degrading accuracy [2]. Our work [2] outlined the method's main ideas, but neither completed the method nor conducted a computer coding implementation. Without additional optimization and refinement, the method will not perform satisfactorily. Also, it remains to be seen how effective the method is. The purpose of this paper is to fill these gaps.

In this paper, we optimize and complete the first general automatic method for explaining machine learning prediction results with no accuracy loss so that it can achieve better performance, present its first computer coding implementation, and demonstrate it on predicting type 2 diabetes diagnosis within the next year. Type 2 diabetes affected 28 million (9%) Americans in 2012 [13]. The average lifetime total direct healthcare cost of treating type 2 diabetes and its complications is around $85,200 [14]. Effective preventive interventions such as lifestyle modification [15-17] and pharmacotherapy [17, 18] exist for type 2 diabetes. Predictive models for type 2 diabetes diagnoses can help clinicians target preventive interventions at the right patients to improve health outcomes and reduce costs [3-6].

**Methods**

*Our previously sketched automatic explanation method*

We first review the draft automatic explanation method for machine learning prediction results outlined in our paper [2]. Our presentation is more general than that in our paper [2], which focuses on asthma outcomes. Our presentation also includes many additional details, insights, and rationales beyond those described in our paper [2]. We will optimize and complete the method in the "Optimizations and refinement" section.

Our automatic explanation method is suitable for integration into predictive modeling tools for various healthcare applications. These tools' users are typically clinicians. Our method works for both categorical and continuous outcome (dependent) variables. For a continuous outcome variable, the method gives explanations for a categorized version of it, such as whether it is above a predetermined threshold value. For ease of presentation, our description focuses on the case of a categorical outcome variable and that every patient has a single data instance (row). The case of a continuous outcome variable or one in which a patient may have multiple data instances can be handled in a similar way.

*Key idea*

Given a predictive modeling problem for a healthcare application, it is often difficult to build an accurate machine learning predictive model that is also easy to interpret. The key idea of our automatic explanation method is to separate explanation from prediction by using two models simultaneously, one for prediction and another for explanation. The first model is used solely for making predictions and aims at maximizing accuracy without being concerned about interpretability. It can be any machine learning model and arbitrarily complex. The second model is a rule-based associative classifier [19-22] used solely for explaining the first model's results without being concerned about its own accuracy.

*Associative classifier*

The associative classifier can handle both categorical and continuous features, a.k.a. input or independent variables. Before being used, continuous features are first converted into categorical ones through discretization [19], e.g., using the minimum description length principle [23]. The associative classifier includes multiple class-based association rules that are easy to interpret. These rules are learned directly from historical data using existing data mining techniques [19-22] rather than obtained from the first model. We optimize these techniques to make them more effective for giving explanations.

Each association rule includes a feature pattern associated with a value $v$ of the outcome variable and is of the form: $p_1$ AND $p_2$ AND … AND $p_m \rightarrow v$. The values of $m$ and $v$ can vary across differing rules. Each item $p_j$ ($1 \leq j \leq m$) is a feature-value pair ($f$, $w$) specifying that feature $f$ takes value $w$ (if $w$ is a value) or a value within $w$ (if $w$ is a range). The rule indicates that a patient's outcome variable is likely to take value $v$ if the patient satisfies $p_1$, $p_2$, …, and $p_m$. An example rule is: the patient had prescriptions of angiotensin-converting-enzyme (ACE) inhibitor in the past three years AND the patient's maximum body mass index recorded in the past three years is $\geq 35 \rightarrow$ the patient will have type 2 diabetes diagnosis within the next year. ACE inhibitor is used mainly for treating hypertension and congestive heart failure.

For a specific association rule $R$, the percentage of patients fulfilling $R$'s left hand side and whose outcome variables take the value on $R$'s right hand side indicates $R$'s coverage and is called $R$'s *support*. Among all patients fulfilling $R$'s left hand side, the percentage of patients whose outcome variables take the value on $R$'s right hand side indicates $R$'s accuracy and is called $R$'s *confidence*. An associative classifier contains association rules at a predetermined level of minimum support (e.g., 1%) and confidence (e.g., 50%).

*Overview of giving explanations*

The outcome variable can take two or more values. Typically, only some of these values are *interesting* ones in the sense that they are of particular interest to the users of the predictive modeling tool for the healthcare application. Explanations are needed for the prediction results matching them. The other values are uninteresting ones and represent normal or safe cases requiring no special attention. No explanation is needed for the prediction results matching them. For example, for predicting type 2 diabetes diagnosis within the next year, having this diagnosis is an interesting value. Not having this diagnosis is an uninteresting value.

For each patient whose outcome variable is predicted by the first model to be of an interesting value, the associative classifier will provide zero or more rules. Each rule lists a reason why the patient's outcome variable is predicted to be of that value. Since some patients' outcome variables can take an interesting value for uncommon reasons that are difficult to pinpoint, the associative classifier makes no guarantee that one or more rules will be displayed for each patient whose outcome variable is predicted to be of an interesting value. Rather, the associative classifier focuses on common reasons, which are more relevant and important for the whole patient population than uncommon ones.

A traditional associative classifier typically achieves lower prediction accuracy than the first model. If used for prediction, the associative classifier could provide incorrect prediction results for some patients whose outcome variables are correctly predicted by the first model to be of an interesting value. A fundamental reason is that such a patient often satisfies the left hand sides of multiple association rules with different values on the right hand sides. The associative classifier would supply an incorrect prediction result if wrong rules for the patient were applied. In comparison, giving explanations for provided prediction results is easier than making correct predictions. The associative classifier can explain the first model's predicted value as long as at least one of the rules whose left hand sides are satisfied by the patient has this value on the right hand side. Hence, in the case that the outcome variable can take an interesting value (positive instance) and an uninteresting value (negative instance), the percentage of patients for whom the associative classifier can explain the first model's correctly predicted interesting value can be higher than the associative classifier's sensitivity.

Our associative classifier is different from traditional ones and includes only association rules for the interesting values of the outcome variable, as it needs to only explain the prediction results matching these values. Consequently, our associative classifier cannot be used to make predictions for all possible values of the outcome variable. In comparison, traditional associative classifiers are used for making predictions and include association rules for all possible values of the outcome variable [19-22].

*Automatically pruning association rules*

Many existing methods for building associative classifiers use a common technique to remove redundant association rules: eliminating more specific and lower-confidence rules in the presence of more general and higher-confidence ones [20]. More specifically, if two rules have the same value on their right hand sides, the items on the left hand side of the first rule are a subset of those on the left hand side of the second rule, and the second rule has lower confidence than the first rule, then the second rule is regarded as redundant and removed.

Usually, a large number of association rules are learned from historical data even after removing redundant rules [19-21, 24]. To avoid overwhelming the users of the predictive modeling tool for the healthcare application, our previous paper [2] uses three additional techniques to further prune rules. First, only features that the first model uses to make predictions are considered in rule construction. Second, all features that the first model uses to make predictions along with all their possible values or value ranges that may appear in the associative classifier [19] are displayed to the designer of the predictive modeling tool for the healthcare application. The designer is usually a clinician and can specify for a feature, which values or value ranges may potentially correlate with the interesting values of the outcome variable in a clinically meaningful way [11, 25]. The other values or value ranges are not allowed to appear in rules. Third, only rules whose left hand sides contain no more than a fixed small number of items (e.g., 4) are considered, as long rules are difficult to comprehend and act on [19].

*Manually refining association rules*

After automatic pruning, all remaining association rules are displayed to the designer of the predictive modeling tool for the healthcare application. The designer examines these rules and removes those that do not make much clinical sense. For each remaining rule, the designer compiles zero or more interventions directed at the reason that the rule shows. For instance, for patients at high risk for type 2 diabetes diagnoses, an example intervention is to enroll an obese patient in a weight loss program. If desired, the designer can differentiate between interventions targeted directly at an interesting value of the outcome variable and other interventions targeted at one or more items on the rule's left hand side. A rule is termed *actionable* if at least one intervention is compiled for it.

*Displaying association rules at prediction time*

At prediction time, for each patient whose outcome variable is predicted by the first model to be of an interesting value, we find all association rules whose left hand sides are satisfied by the patient. Actionable rules are displayed above nonactionable ones, each in descending order of confidence. Alternatively, nonactionable rules can be omitted. Confidence is listed next to each rule. Frequently, many rules apply to a patient. To avoid overwhelming the user of the predictive modeling tool for the healthcare application, by default no more than a fixed small number $n_r$ of rules (e.g., 5) are shown. If he/she chooses, the user can view all rules applicable to the patient. To help the user find appropriate tailored interventions, interventions associated with an actionable rule are listed next to the rule.

**Optimizations and refinement**

Our previous paper [2] only outlined the main ideas of our automatic explanation method for machine learning prediction results, but did not complete the method. Without additional optimization and refinement, this method will not perform satisfactorily. In this section, we optimize and refine this method to make it complete and achieve better performance.

*Further automatically pruning association rules*

Even if all four techniques mentioned in the "Automatically pruning association rules" section are used for pruning association rules, too many rules often still remain and overwhelm the users of the predictive modeling tool for the healthcare application. Since our associative classifier is used for giving explanations rather than for making predictions, we can use this property to further automatically prune rules. Consider two rules with the same value on their right hand sides. The items on the left hand side of the first rule are a subset of those on the left hand side of the second rule. Also, the first rule's confidence is lower than the second rule's confidence, but not by more than a predetermined threshold such as 10%. In other words, the first rule is more general and has slightly lower confidence than the second rule. In this case, the second rule is non-redundant for making predictions and will not be removed by the first technique described in the "Automatically pruning association rules" section. However, the first rule can give an explanation for every patient covered by the second rule without losing much useful information. Thus, the second rule is redundant for giving explanations and is removed.

*A software widget for helping compile interventions for actionable association rules*

In the process of manually refining association rules, the designer of the predictive modeling tool for the healthcare application will compile interventions for every remaining actionable association rule. This work is labor intensive when the number of such rules is large. We can build a software widget to facilitate and help expedite this work.

We notice that many interventions relate to a value or value range of a single feature, which corresponds to an *actionable* item. An intervention related to a single value or value range is likely relevant for an association rule containing this value or value range on the left hand side. Based on this insight, the widget lists every value or value range of a feature that is allowed to appear in an association rule. For each such value or value range, the designer of the predictive modeling tool for the healthcare application lists zero or more interventions related to it. When compiling interventions for the remaining actionable rules, for each actionable rule the widget automatically displays all listed interventions related to any single value or value range on the rule's left hand side, if any. Then the designer can select among these interventions without having to laboriously dig out these interventions himself/herself.

*Displaying diversified association rules at prediction time*

As mentioned in the "Displaying association rules at prediction time" section, at prediction time usually many association rules apply to a patient, but by default only a fixed small number $n_r$ of them (e.g., 3) are shown to the user of the predictive modeling tool for the healthcare application. The rule sorting method proposed in our previous paper [2] is that actionable rules are displayed above nonactionable ones, each in descending order of confidence. This method is acceptable in the uncommon case that the user chooses to view all rules applicable to the patient, but is sub-optimal in the default common case of displaying the small number $n_r$ of rules due to redundancy among rules. For example, two rules applicable to a patient often differ by only one item and share all of the other items on their left hand sides. After the user sees one rule, the other rule will likely provide little additional information to the user. When the first rule is shown, it can be more informative to display a third rule that shares no items with the first rule on their left hand sides than to display the second rule. In other words, it is desirable to display diversified rules to maximize the amount of information that the user can obtain from the small number $n_r$ of rules. This is similar to the case of information retrieval, where users prefer diversified search results [26-28].

To diversify the small number $n_r$ of association rules for display, we use the following approach to let the $n_r$ rules have completely different items on the left hand sides if possible. We first use the method proposed in our previous paper [2] to sort all rules applicable to the patient. Then the number of actionable rules is checked. If this number is no larger than $n_r$, all actionable rules are displayed. Otherwise, we choose $n_r$ from all actionable rules for display. We select the top ranked actionable rule and then go through the rest of the list of actionable rules one by one. A rule is selected if no (actionable) item on its left hand side appears on the left hand side of any previously selected rule. We stop after $n_r$ rules are selected. If we still have not selected $n_r$ rules when reaching the end of the list, we go through the list again to select whichever unchosen rules we encounter first until $n_r$ rules are chosen. After selecting the actionable rules for display, if there is still room, a similar approach is used to choose the nonactionable rules for display. This case occurs when the number of actionable rules is smaller than $n_r$.

An alternative method for choosing $n_r$ from all actionable rules for display is as follows. The designer of the predictive modeling tool for the healthcare application categorizes all possible actionable items and assigns a positive weight to each category. A larger weight reflects higher clinical importance. Each actionable item is initially given a weight equal to the weight of the category that it falls into. An actionable rule's weight is the sum of the weights of all actionable items on the rule's left hand side. We go through the list of actionable rules $n_r$ times and choose one rule at a time. Each time among the unchosen rules, we select the one with the largest weight that we encounter first. For each actionable item on the rule's left hand side, we set the item's weight to zero. Then we re-compute the weight of each remaining unchosen rule before starting the rule selection process for the next round.

*Computer coding implementation*

We did a computer coding implementation of our method for automatically explaining machine learning prediction results. In our implementation, we required that each association rule contains no more than four items on the left hand side. We set the minimum support threshold to 1% and the minimum confidence threshold to 50%. These threshold values were shown to work well in Liu *et al* [19]. Although it does not seem to be high, the minimum confidence threshold of 50% is already much larger than the typical percentage of patients whose outcome variables have a specific interesting value. For instance, in our demonstration test case, this percentage is 19%.

*Demonstration test case*

Our automatic explanation method works for any predictive modeling problem where machine learning is used. As a demonstration test case, we evaluated our automatic explanation method on predicting type 2 diabetes diagnosis within the next year. We used the electronic medical record data set from the Practice Fusion diabetes classification competition [29]. The data set is de-identified and publicly available. It contains both historical 3-year records and the following year's labels of 9,948 patients from all 50 states in the United States. Among these patients, 1,904 were diagnosed with type 2 diabetes within the next year. The data set includes information on patient demographics, diagnoses, allergies, immunizations, lab results, medications, smoking status, and vital signs.

The champion machine learning predictive model of the Practice Fusion diabetes classification competition was formed by combining/stacking eight boosted regression trees and four random forests using a generalized additive model with cubic splines [29]. The model is complex and non-interpretable. In our demonstration, our automatic explanation method was used to explain the model's results. We did not attempt to modify the model and improve its accuracy, as this work's goal is to show the feasibility of automatically explaining results for any machine learning predictive model without degrading accuracy, rather than to achieve maximum possible accuracy for a specific predictive modeling problem.

*Experimental approach*

The champion machine learning predictive model computes each patient's probability of having type 2 diabetes diagnosis within the next year. We used the standard Youden's index [30] to select the optimum cut-off point for the computed probability for deciding whether the patient will have type 2 diabetes diagnosis within the next year. This approach is equivalent to maximizing the sum of sensitivity and specificity. Our automatic explanation method was used to give explanations for the interesting value of the outcome variable: the patient will have type 2 diabetes diagnosis within the next year.

We randomly chose 80% of patients to form the training set to train both the champion machine learning predictive model and the associative classifier. We used the other 20% of patients to form the test set to evaluate the performance of the champion machine learning predictive model and our automatic explanation method.

An internal medicine doctor who is also a diabetologist participated in this study as a clinical expert to provide expertise on type 2 diabetes. He served several parts of the role of the designer of the predictive modeling tool for the healthcare application if there were such a designer and tool.

*Ethics approval*

This study used publicly available, de-identified data and is regarded as non-human subject research by the Institutional Review Board of the University of Utah.

**Results and discussion**

*Performance of the champion machine learning predictive model*

On the test set, the champion machine learning predictive model achieved reasonable performance: an area under the receiver operating characteristic curve of 0.884, an accuracy of 77.6%, a sensitivity of 84.4%, a specificity of 76.1%, a positive predictive value of 44.8%, and a negative predictive value of 95.5%.

*The number of remaining association rules at different stages of the rule pruning process*

The associative classifier was trained on the training set. By considering only features that the champion machine learning model used to make predictions in rule construction, 178,814 association rules were obtained. After eliminating more specific and lower-confidence rules in the presence of more general and higher-confidence ones [20], 16,193 rules remained.

Fig. 1 shows the number of remaining association rules vs. the confidence difference threshold when the technique described in the "Further automatically pruning association rules" section was used for further pruning rules. This technique removed more specific rules in the presence of more general ones with slightly lower confidence. Initially when the confidence difference threshold is small, the number of remaining rules drops rapidly as the confidence difference threshold increases. Once the confidence difference threshold becomes 0.1 or larger, the number of remaining rules no longer drops much any more. In our computer coding implementation, we chose the confidence difference threshold to be 0.1, which struck a good balance between sufficiently reducing the number of rules and keeping high-confidence rules. 2,601 rules remained at the end of this stage of the rule pruning process.

Our clinical expert specified the values or value ranges of features that may potentially correlate with the interesting value of the outcome variable in a clinically meaningful way. By eliminating association rules containing any other values or value ranges, 415 rules remained. Overall, our automatic rule pruning techniques reduced the number of rules by three orders of magnitude and made it feasible for a human expert to manually refine the remaining rules.

Each association rule lists a reason why a patient is predicted to have type 2 diabetes diagnosis within the next year. Our clinical expert examined the 415 rules and regarded all of them as clinically meaningful, mainly because the rule pruning process allowed only those values or value ranges of features that may potentially correlate with the interesting value of the outcome variable in a clinically meaningful way to appear in these rules. These rules were all kept in the final associative classifier. 283 of these rules were actionable.

For external validation of the 415 rules, two other doctors who are familiar with type 2 diabetes independently examined these rules. These two doctors regarded 97.8% and 98.3% of these rules as clinically meaningful, respectively. The review results of these two doctors achieved good agreement (kappa=0.745).

*Example rules in the associative classifier*

To give the reader a feeling of the rules in the associative classifier, we list five example rules as follows.

Rule 1: the patient had prescriptions of angiotensin-converting-enzyme (ACE) inhibitor in the past three years AND the patient's maximum body mass index recorded in the past three years is $\geq 35$ $\rightarrow$ the patient will have type 2 diabetes diagnosis within the next year. ACE inhibitor is used mainly for treating hypertension and congestive heart failure. Obesity, hypertension, and congestive heart failure are known to correlate with type 2 diabetes. One intervention compiled for the rule is to enroll the patient in a weight loss program.

Rule 2: the patient had prescriptions of loop diuretics in the past three years AND the patient had $\geq 23$ diagnoses in total in the past three years $\rightarrow$ the patient will have type 2 diabetes diagnosis within the next year. Loop diuretics are used for treating hypertension. Hypertension and having a large number of diagnoses are known to correlate with type 2 diabetes.

Rule 3: the patient had a diagnosis of coronary atherosclerosis disease in the past three years AND the patient is over 65 AND the patient had $\geq 15$ diagnoses per year in the past three years $\rightarrow$ the patient will have type 2 diabetes diagnosis within the next year. Coronary atherosclerosis disease, old age, and having a large number of diagnoses are known to correlate with type 2 diabetes.

Rule 4: the patient had ≥ 6 diagnoses of hyperlipidemia in the past three years AND the patient had prescriptions of statins in the past three years AND the patient had ≥ 9 prescriptions in the past three years → the patient will have type 2 diabetes diagnosis within the next year. Hyperlipidemia refers to high lipid (fat) level in the blood. Statins are used for lowering cholesterol. Hyperlipidemia, high cholesterol level, and using many medications are known to correlate with type 2 diabetes.

Rule 5: the patient had ≥ 5 diagnoses of hypertension in the past three years AND the patient had prescriptions of statins in the past three years AND the patient had ≥ 11 doctor visits in the past three years → the patient will have type 2 diabetes diagnosis within the next year. Hypertension, high cholesterol level, and frequent doctor visits are known to correlate with type 2 diabetes. One intervention compiled for the rule is to suggest the patient to make lifestyle changes to help lower his/her blood pressure.

*Performance of our automatic explanation method*

Our automatic explanation method was evaluated on the test set. For the champion machine learning model, our automatic explanation method explained its prediction results for 87.4% of patients who were correctly predicted by it to have type 2 diabetes diagnosis within the next year. This percentage is sufficiently high for our method to be used in routine clinical practice. For each such patient, our method on average gave 41.7 explanations, of which 29.8 were actionable. Each explanation corresponded to one rule. When explanations were limited to actionable ones, our automatic explanation method could still explain the model's prediction results for 86.1% of patients who were correctly predicted by the model to have type 2 diabetes diagnosis within the next year.

As is typical with association rule mining, much redundancy exists among the rules in the associative classifier. Many rules differ from each other only slightly, e.g., by a single item on the left hand side. The same comment holds for all of the rules applicable to a typical patient. Hence, although many rules can apply to a patient, the total amount of information contained in these rules is usually much smaller. For example, consider the number of identified actionable items of a patient, which is defined as the total number of distinct actionable items in all of the rules applicable to the patient. For each patient for whom our automatic explanation method could explain the champion machine learning model's correct prediction result of having type 2 diabetes diagnosis within the next year, the number of identified actionable items of the patient was on average 3.5. This number is much smaller than 29.8, the average number of actionable rules applicable to such a patient.

Fig. 2 shows the distribution of patients by the number of association rules applicable to a patient. This distribution is heavily skewed towards the left and has a long tail. As the number of rules applicable to a patient increases, the number of patients, each of whom is covered by this number of rules, generally tends to decrease, although this decrease is non-monotonic. The maximum number of rules applicable to a single patient is rather large: 210, although only one patient is covered by this number of rules.

Fig. 3 shows the distribution of patients by the number of actionable rules applicable to a patient. This distribution is similar to that shown in Fig. 2. The maximum number of actionable rules applicable to a single patient is rather large: 133, although only one patient is covered by this number of actionable rules.

Fig. 4 shows the distribution of patients by the number of identified actionable items of a patient. The maximum number of identified actionable items of a single patient is 8, much smaller than the maximum number of (actionable) rules applicable to a single patient.

Our automatic explanation method could give explanations for 80.3% of patients who will have type 2 diabetes diagnosis within the next year. This percentage is lower than the success rate (87.4%) for our automatic explanation method to explain the champion machine learning model's correct prediction results of having type 2 diabetes diagnosis within the next year. One possible reason is that the results of different models are correlated rather than are completely independent of each other. Among patients who will have type 2 diabetes diagnosis within the next year, some represent hard cases that are difficult for any model to correctly predict or explain their outcomes. Similarly, among patients who were correctly predicted by the champion machine learning model to have type 2 diabetes diagnosis within the next year, many represent easy cases that are also not difficult for other models to correctly predict or explain their outcomes.

## Conclusions

We optimized and completed the first general automatic method for explaining machine learning prediction results with no accuracy loss. Our demonstration on predicting type 2 diabetes diagnosis showed that it is feasible to automatically explain results for any machine learning predictive model without degrading accuracy. Future studies will demonstrate our method on other predictive modeling problems in healthcare.

## Competing interests

The author declares that he has no competing interests.

## Authors' contributions

GL conceptualized and conducted the study, drafted the manuscript, and read and approved the final manuscript.

**Authors' information**

Department of Biomedical Informatics, University of Utah, Suite 140, 421 Wakara Way, Salt Lake City, UT 84108, USA

**Acknowledgments**

We thank Prasad Unni for providing his clinical expertise on type 2 diabetes, and Chandan Reddy, Katherine Sward, Zhongmin Wang, and Philip J. Brewster for helpful discussions.

This is a pre-print of an article published in Health Information Science and Systems. The final authenticated version is available online at: https://doi.org/10.1186/s13755-016-0015-4.

[27] Luo G, Tang C, Yang H, Wei X. MedSearch: a specialized search engine for medical information retrieval. *Proc. CIKM* 2008:143-52.

[28] Santos RLT, Macdonald C, Ounis I. Search result diversification. *Foundations and Trends in Information Retrieval* 2015;9(1):1-90.

[29] Practice Fusion diabetes classification homepage. https://www.kaggle.com/c/pf2012-diabetes, 2016.

[30] Youden's J statistic. https://en.wikipedia.org/wiki/Youden%27s_J_statistic, 2016.

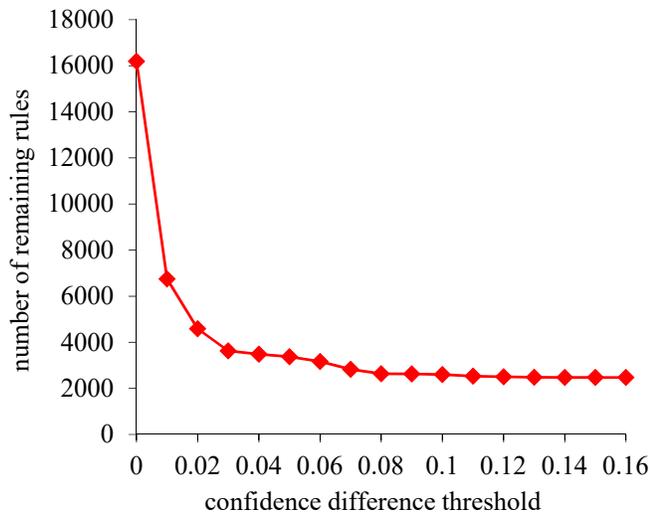

**Figure 1**. The number of remaining association rules vs. the confidence difference threshold.

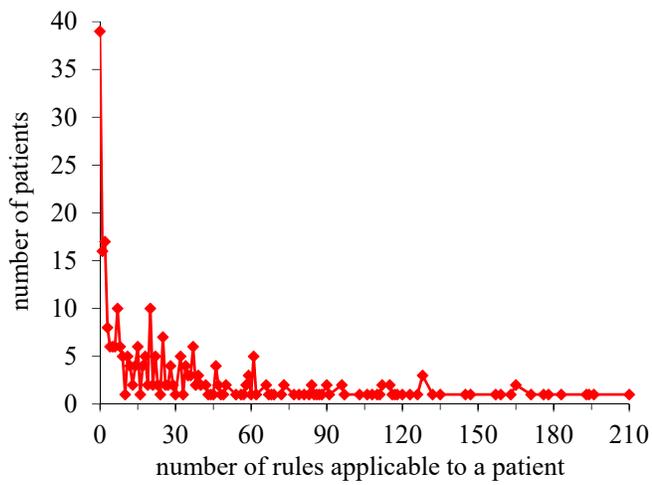

**Figure 2**. Distribution of patients by the number of association rules applicable to a patient.

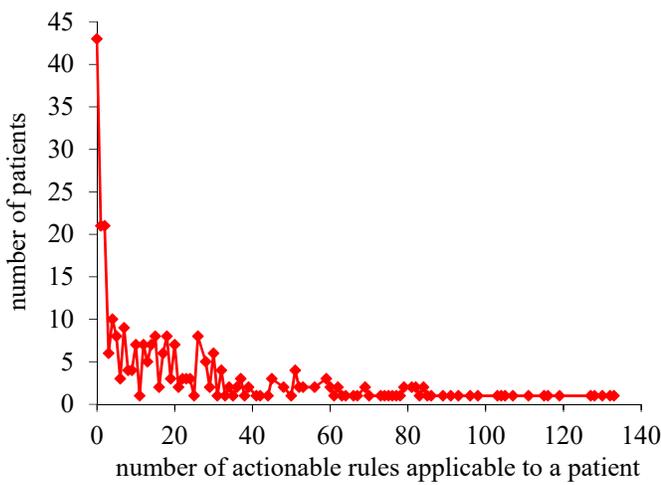

**Figure 3**. Distribution of patients by the number of actionable rules applicable to a patient.

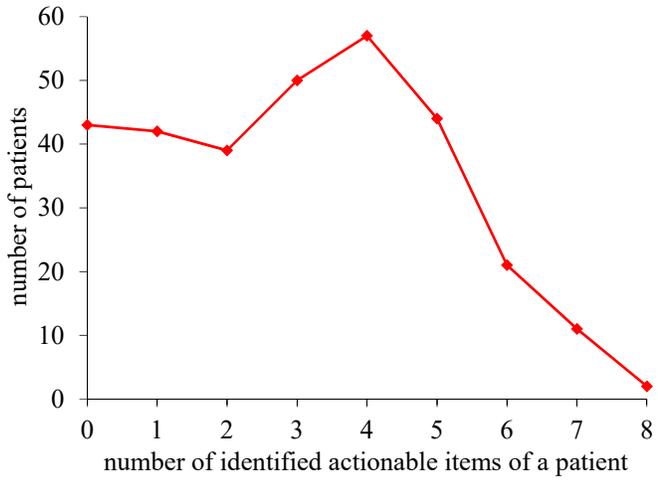

**Figure 4**. Distribution of patients by the number of identified actionable items of a patient.